\documentclass[a4paper]{article}

\usepackage[utf8]{inputenc}
\usepackage{graphicx}
\usepackage{amsmath}

\newcommand{\X}{\ensuremath{\mathbf{X}}}
\newcommand{\V}{\ensuremath{\mathbf{V}}}
\newcommand{\Y}{\ensuremath{\mathbf{Y}}}

\renewcommand{\H}{\ensuremath{\mathbf{H}}}

\newcommand{\x}{\ensuremath{\mathbf{x}}}
\newcommand{\h}{\ensuremath{\mathbf{h}}}
\renewcommand{\v}{\ensuremath{\mathbf{v}}}
\newcommand{\y}{\ensuremath{\mathbf{y}}}
\renewcommand{\b}{\ensuremath{\boldsymbol{\beta}}}

\begin{document}

\title{Incremental ELMVIS for unsupervised learning}

\author{
Anton Akusok$^1$
\and Emil Eirola$^1$ 
\and Yoan Miche$^{2,3}$ 
\and Ian Oliver$^2$
\and Kaj-Mikael Bj\"{o}rk$^4$
\and Andrey Gritsenko$^{5,6}$
\and Stephen Baek$^5$
\and Amaury Lendasse$^{1,5,6}$
}

\date{
    $^1$ Arcada University of Applied Sciences, Helsinki, Finland \\
    $^2$ Nokia Solutions and Networks Group, Espoo, Finland \\
    $^3$ Department of Computer Science, Aalto University, Finland \\
    $^4$ Risklab at Arcada University of Applied Sciences, Finland \\
    $^5$ Department of Mechanical and Industrial Engineering, \\The University of Iowa, Iowa City, USA \\
    $^6$ The Iowa Informatics Initiative,\\ The University of Iowa, Iowa City, USA
}

\maketitle

\begin{abstract}

An incremental version of the ELMVIS+ method is proposed in this paper. It iteratively selects a few best fitting data samples from a large pool, and adds them to the model. The method keeps high speed of ELMVIS+ while allowing for much larger possible sample pools due to lower memory requirements. The extension is useful for reaching a better local optimum with greedy optimization of ELMVIS, and the data structure can be specified in semi-supervised optimization. The major new application of incremental ELMVIS is not to visualization, but to a general dataset processing. The method is capable of learning dependencies from non-organized unsupervised data -- either reconstructing a shuffled dataset, or learning dependencies in complex high-dimensional space. The results are interesting and promising, although there is space for improvements.

\end{abstract}

\section{Introduction}

The ELMVIS method~\cite{cambria_extreme_2013} is an interesting Machine Learning method that optimize a cost function by changing assignment between two sets of samples, or by changing the order of samples in one set which is the same. The cost function is learned by an Extreme Learning Machine (ELM)~\cite{huang_extreme_2006,huang_extreme_2012,huang_what_2015}, a fast method for training feed-forward neural networks with convenient mathematical properties~\cite{huang_universal_2006,miche_opelm_2010}. Such optimization problem is found in various applications like open-loop Traveling Salesman problem~\cite{gutin_traveling_2002} or clustering~\cite{alpaydin_introduction_2014} (mapping between samples and clusters), but not in Neural Networks. ELMVIS is unique in a sense that it combines the optimal assignment task with neural network optimization problem; the latter is optimized at each step of ELMVIS.

A recent advance in ELMVIS+ method~\cite{akusok_elmvis_2016} set its runtime speed comparable or faster than other state-of-the-art methods in visualization application. However there are unresolved problems like a greedy optimization leading to a local optimum. Also ELMVIS+ can be applied to a much wider range of problems than a simple visualization or a visualization accounting for the class information~\cite{gritsenko_combined_2016}, which have not been tested or reported yet. This paper addresses the aforementioned drawbacks, and presents the most recent research advances in the family of ELMVIS methods.

The proposed incremental ELMVIS allows for iterative growth of dataset size and model complexity. Incremental ELMVIS learns an approximate global data structure with a few data samples and a simple ELM model, because at a very small scale global and local optimums are similar or the same. Then more data samples are added to the model, choosing the ones that better fit an existing ELM. After adding a batch of new samples, the current model is refined by ELMVIS+ method. This refinement keeps the global optimum due to the greedy optimization and only small changes. More neurons are added to ELM as the dataset size grows, to better separate the data.

Iterative ELMVIS is useful for semi-supervised learning, starting from the data with known outputs and adding more data with unknown outputs, simultaneously updating the model. It can even apply to completely unsupervised datasets, where it finds an input-output dependency, learns it with ELM model, and then simultaneously expands the supervised part of a dataset and updates an ELM model that encodes the input-output dependency.

The experiments have shown the ability of iterative ELMVIS to improve global optimum, successfully perform semi-supervised and unsupervised learning with complex tasks. Current version of the method is found limited to good separation between only two classes in data (which it learns first), ignoring samples of the additional classes until the first two ones are exhausted, and poorly fitting the additional classes into the learned two-class model. Solution to this problem will be considered in further works on the topic.

\section{Methodology}

An iterative extension of ELMVIS+ methodology is rather straight-forward, as explained below. ELMVIS methods start with a list of (visualization) samples and an unordered set of data samples; and it finds an optimal order of data samples in the set by a greedy search with changing positions of many samples at once (ELMVIS) or only two samples at once (ELMVIS+). 

Iterative ELMVIS splits data samples into fixed, candidate and available ones. Fixed samples have their order fixed and cannot be moved by an iterative ELMVIS. Candidate samples are a small number of samples which are chosen from candidate+available ones to maximize the cost function. This cost function takes into account fixed and current candidate samples, but ignores the available samples. Once current candidate samples are chosen optimally, they are added to the fixed ones, and the method is repeated with a few more candidate samples - until the available data samples are exhausted.

\subsection{Extreme Learning Machine}

Extreme Learning Machine is a way of training feedforward neural networks~\cite{haykin_neural_1998} with a single hidden layer that features randomly assigned input weights~\cite{huang_insight_2014}, explicit non-iterative solution for output weights and an extreme computation speed and scalability~\cite{akusok_highperformance_2015}. This model is used as a non-linear cost function in all ELMVIS methods. This short summary introduces the notations to the reader.

The goal of ELM model is to approximate the projection function $\hat{\y}_i \approx f(\x_i)$ using a representative training dataset. As ELM is a deterministic model, the function $f()$ is assumed to be deterministic, and a noise $\epsilon$ is added to cover the deviation of true outputs $\y$ from the predictions by a deterministic function $f()$
\begin{equation}
    \y = f(\x) + \epsilon
\end{equation}

The Extreme Learning Machine~\cite{huang_extreme_2006} (ELM) is a neural network with $d$ input, $L$ hidden and $c$ output neurons. The hidden layer weights $\mathbf{W}_{d \times L}$ and biases $\text{bias}_{1 \times L}$ are initialized randomly and are fixed. The hidden layer neurons apply a transformation function $\phi$ to their outputs that is usually a non-linear transformation function with bounded output like sigmoid or hyperbolic tangent.

The output of the hidden layer is denoted by $\h$ with an expression
\begin{equation}
    \h_i = \phi(\x_i\mathbf{W} + \text{bias})
\end{equation}
where the function $\phi()$ is applied element-wise, and can also be gathered in a matrix $\H_{N \times L}$ for convenience.

The output layer of ELM poses a linear problem $\H\b = \Y$ with unknown output weights $\b_{L \times c}$. The solution is derived from the minimization of the sum of squared residuals $\mathbf{r}^2 = (\y - \hat{\y})^2$, that gives the ordinary least squares solution $\b = \H^\dagger \Y$, where $\H^\dagger$ is a Moore-Penrose pseudoinverse~\cite{rao_generalized_1972} of the matrix $\H$.

\subsection{ELMVIS+ method}

ELMVIS+ method~\cite{akusok_elmvisplus_2016} approximates a relation between visualization (or input) space $\mathcal{V}$ and data space $\mathcal{X}$ by an ELM model. The task has $N$ representative samples $\v \in \mathcal{V}$ and $\x \in \mathcal{X}$, but their order is unknown. The method assumes fixed order of samples $\v$ joined in matrix $\V$, and finds a suitable order of samples $\x$ joined in matrix $\X$ by exchanging pairs of rows in $\X$. Contrary to a common use of ELM, data samples $\x$ are the outputs of ELM and visualization coordinates $\v$ are the inputs (thus ELM predicts original data $\hat{\x}$). Visualization coordinates $\V$ are chosen arbitrary and fixed -- they can be distributed randomly with normal or uniform distribution, or initialized on a regular grid. 

The optimization criterion is a cosine similarity between $\X$ and $\hat{\X}$ predicted by ELM. A low error means that its possible to reconstruct data from the given visualization points, thus the visualization points keep information about the data. The reconstruction is approximated by the ELM model in ELMVIS.

There is an explicit formula for a change of error (negative cosine similarity) for swapping two rows in $\X$ and re-training ELM with this new dataset. The readers can refer to the original paper~\cite{akusok_elmvisplus_2016} for the full formula. It is based on the expression for the change on error $\Delta_E$ in case a row $\x_a$ in $\X$ is changed by $\delta \in \mathcal{X}$ amount:

\begin{eqnarray}
\Delta_E &=& \sum_{j=1}^{d} \Big( \mathbf{A}_{a,a} \delta_j^2 + 2 \hat{\x}_{a,j} \delta_j \Big) \label{eq:delta}\\ 
\hat{\X} &\leftarrow& \hat{\X} - \mathbf{A}_{:,a} \times \delta \label{eq:update1}\\
\X_{:,a} &\leftarrow& \X_{:,a} + \delta \label{eq:update2}
\end{eqnarray}

\subsection{Incremental ELMVIS}

Incremental ELMVIS splits all data samples in three groups: fixed, candidate and available samples. The separation is maintained with two indexes: $i_A$ is the number of fixed samples, and $i_B$ is the number of fixed+candidate ones.

Incremental ELMVIS works similar to ELMVIS+. First, the initial numbers of fixed and candidate samples are given by $i_A$ and $i_B$. Then swap indexes $a \in [ i_{A}, i_{B} ]$ and $b \in [ i_{B}+1, N]$ are selected randomly to replace one candidate sample with an available one. The change of error $\Delta_E$ is computed by the formula~\eqref{eq:delta} for the change in the candidate row of the data matrix $\X$. Compared to ELMVIS+, the change in the available sample is ignored. Then for the negative $\Delta_E$, an update step is performed for sample $\x_a$ as in equation~\eqref{eq:update1} and for both $\x_a$ and $\x_b$ as in equation~\eqref{eq:update2}. 

If there is no improvement during a large number of swaps, the current candidate samples are added to the fixed ones ($i_A \leftarrow i_B$), and $k$ more samples are added as candidates ($i_B \leftarrow i_B + k$). Candidate samples are already initialized with the data samples at indexes $\x_{i_A}, \ldots, \x_{i_B}$. The method then repeats for another iteration. Iterations stop when no available samples are left.

In the original ELMVIS+, matrix $\mathbf{A}$ took the most space and limited the maximum amount of processed samples (its memory size is $\mathcal{O}(N^2)$). In incremental ELMVIS, only a $\mathbf{A}_{i_B \times i_B}$ part of the whole matrix $\mathbf{A}$ is needed. That relaxed memory requirements of the method, and allows to use a very large pool of available samples. The memory constraints of incremental ELMVIS apply only to the number of optimized data samples.

\section{Experimental Results}

Incremental ELMVIS method is developed for two main applications. The first one is achieving a better global optimum in ELMVIS+. The original EMLVIS+ is a good visualization method, however it has an unwanted feature: with a large number of neurons in ELM it fragments clusters in the visualized data. This happens with large amount of data and a complex ELM model. An iterative ELMVIS that starts with small amount of data and a simple model keeps all similar data together; then more data samples are gradually added while the total picture changes little due to local minimum in ELMVIS+ optimization.

The second application is finding unknown relations in datasets. This is an unsupervised learning field relevant to the current Big Data trends, when a large amount of interesting data is available - but there pre-processing like manual labeling or classification. It is possible to extract relations inside data automatically by iteratively growing an ELMVIS+ model between two sets of data samples (they don't have to be related to visualization). Results for both applications are presented below.

\subsection{Better Optimum with ELMVIS}
\label{sec:evbasic}

ELMVIS+ method is fast and works with large datasets, but it has a greedy optimization approach that leads to local optimality of the solution. Such local optimum is close to a global one for small datasets and simple ELM models, but with a large dataset and many neurons in ELM model the visualization data is split into multiple small clusters with local optimality, non-representative of a global picture. 

\begin{figure}
	\centering 
	\includegraphics[width=0.49\textwidth]{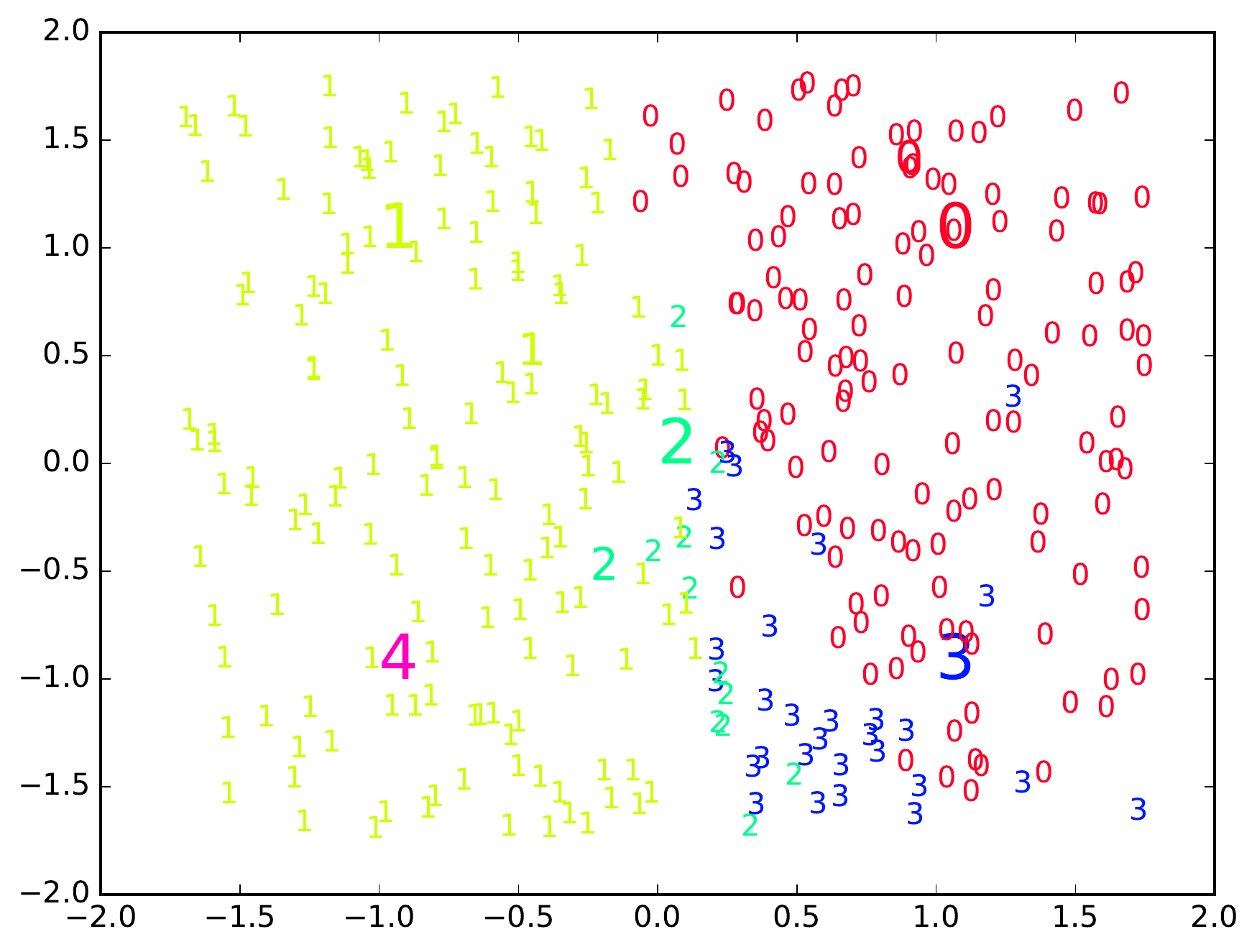}
	\includegraphics[width=0.49\textwidth]{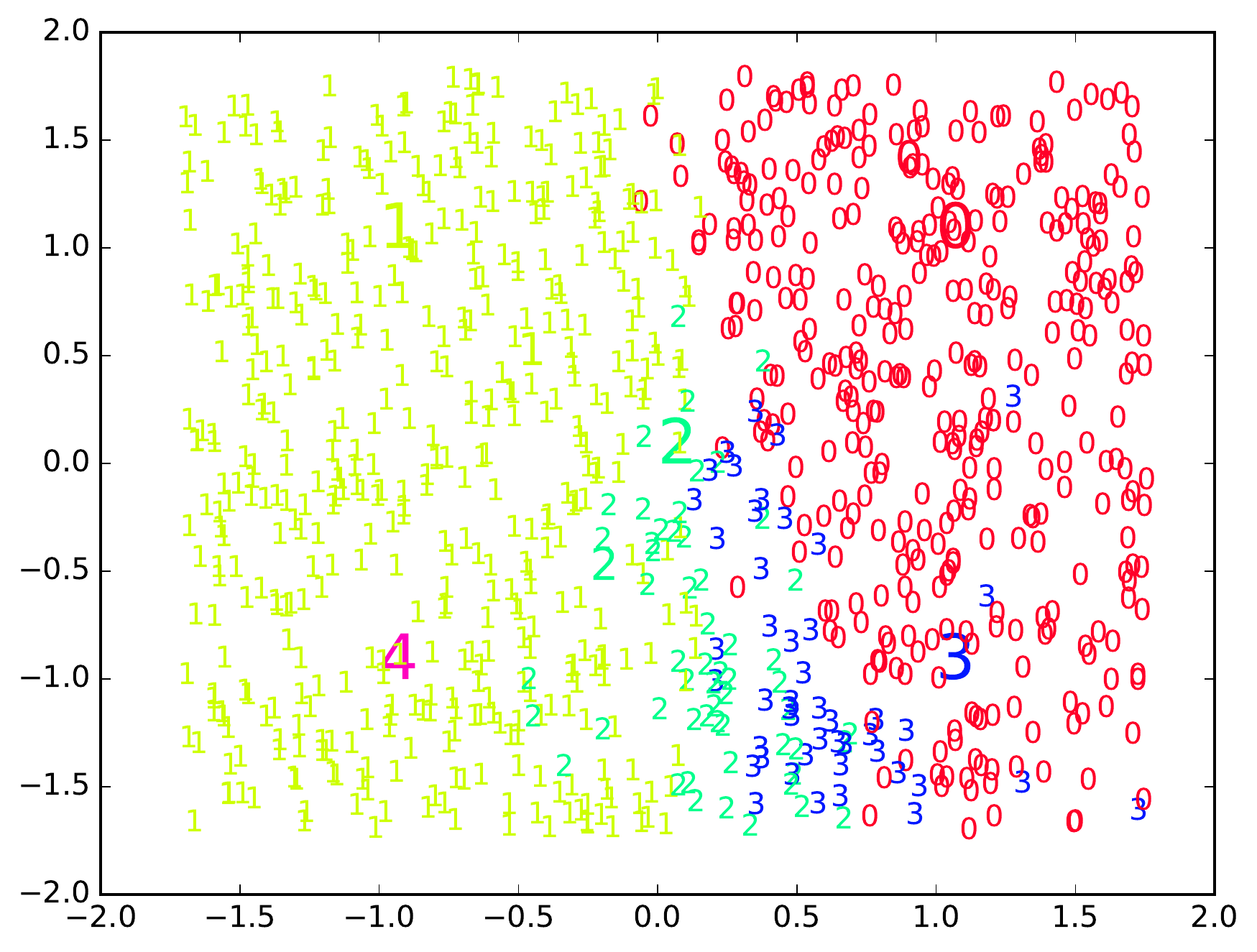}
	\caption{Visualization of 300 (\emph{left}) and 900 (\emph{right}) MNIST digits on a two-dimensional space with incremental ELMVIS. The visualization is initialized with 5 digits shown in larger font. Digit symbols and colors are for the presentation purpose only; ELMVIS method does not have access to them and works with raw pixel data.}
	\label{fig:us1}
\end{figure}

A better global optimality is achievable with an incremental ELMVIS. This experiment uses MNIST digits~\cite{y._lecun_gradient-based_1998} with their original $28 \times 28$ features (grayscale pixels), with 500 digits for each of the classes 0-4. It starts by seeding several cluster as shown on Figure~\ref{fig:us1} by bold samples, and a simple ELM model. Then gradually added data fits into the existing model (Figure~\ref{fig:us1}, left). An ELM learns an easy separation between two clusters, and an incremental ELMVIS prefers to add samples of these clusters until they are available (Figure~\ref{fig:us1}, right).

\begin{figure}
	\centering 
	\includegraphics[width=0.49\textwidth]{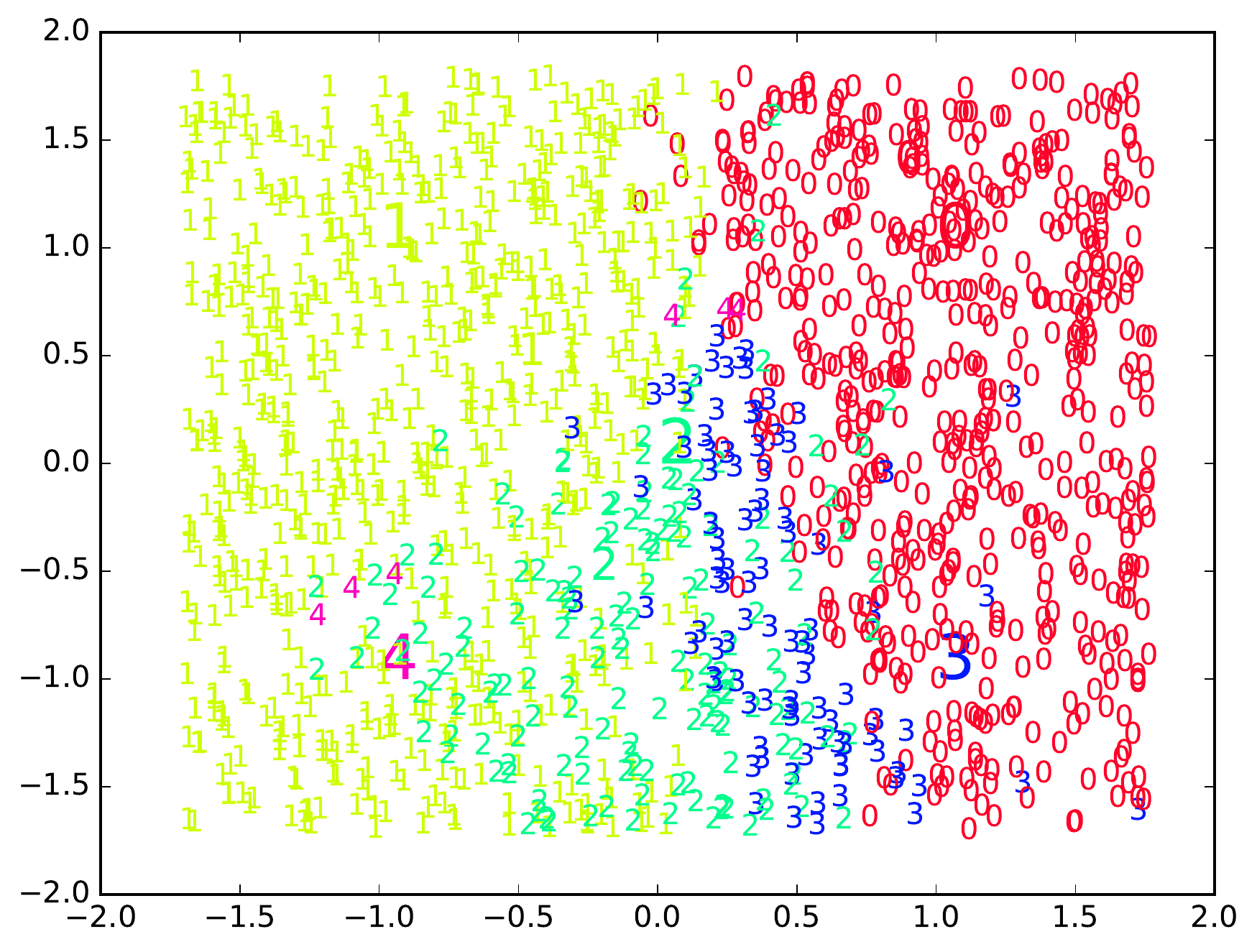}
	\includegraphics[width=0.49\textwidth]{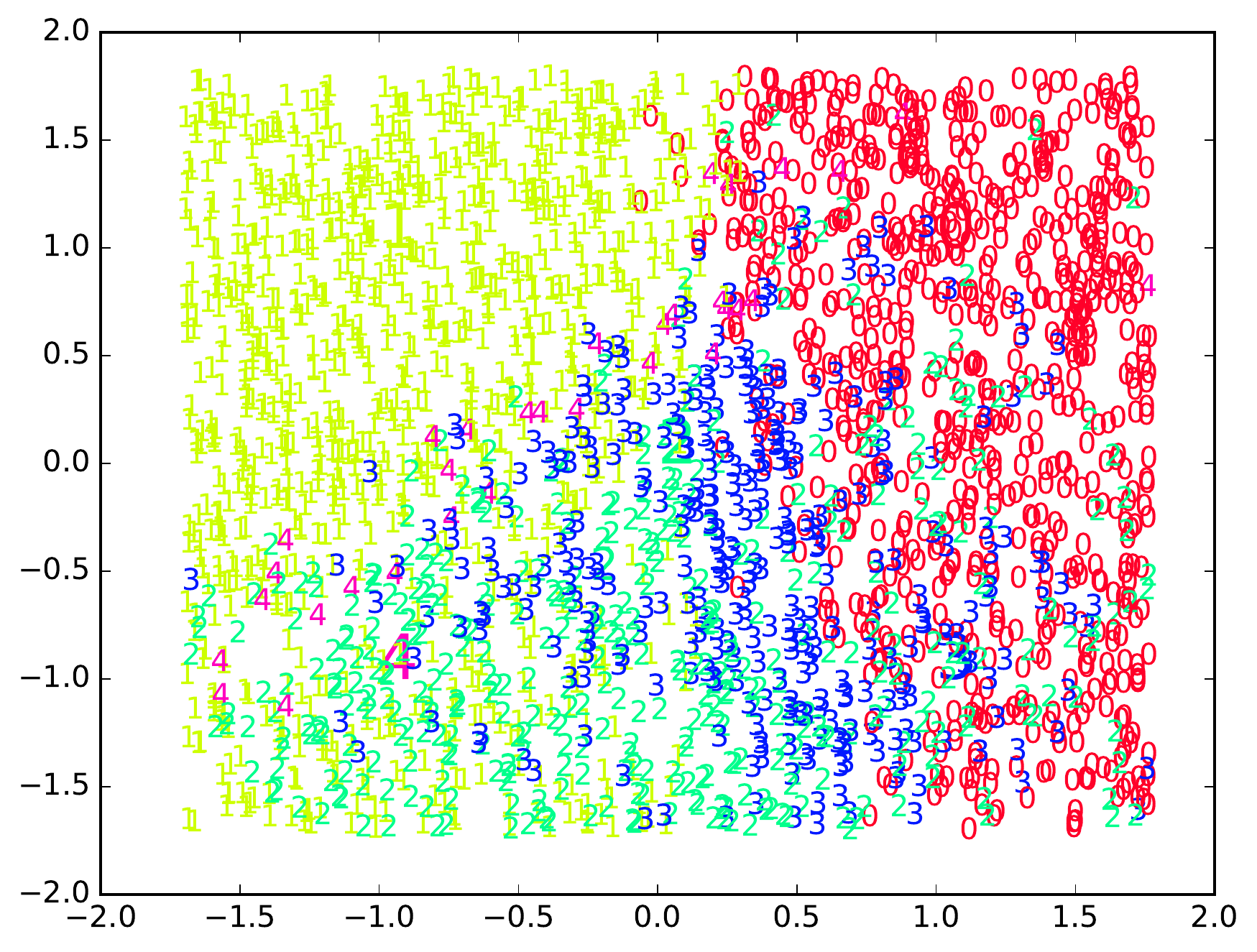}
	\caption{Visualization of 1500 (\emph{left}) and 2500 (\emph{right}) MNIST digits on a two-dimensional space with incremental ELMVIS. There is only 500 samples of each class; ELMVIS has to add more classes then it exhausted zeros and ones in the available set. Digit symbols and colors are for the presentation purpose only; ELMVIS method does not have access to them and works with raw pixel data.}
	\label{fig:us2}
\end{figure}

The incremental ELMVIS learns a good model that separated between two different kinds of data. Then there is no data samples of these two types left, it begins adding more types, starting at the boundary (Figure~\ref{fig:us2}, left). These additional classes are mapped to a single area, although they go over the two previously learned cluster as there is no space left on the visualization (Figure~\ref{fig:us2}, right). The ELMVIS still ignores the last available class (digits 4) because it is not represented on the visualization space.

\subsection{Better Optimum with Semi-supervised ELMVIS}

In the previous experiment, there were no sharp borders between clusters because ELMVIS used all the visualization space to show only two clusters, and then has to map additional data clusters other them. Sharper borders can be obtained by running the original ELMVIS+ on the fixed set of data points after each iteration of the incremental ELMVIS. That will make space for more clusters by compacting the existing ones; while still preserving the global structure as ELMVIS+ optimization goes to the local optimum only.

\begin{figure}
	\centering 
	\includegraphics[width=0.49\textwidth]{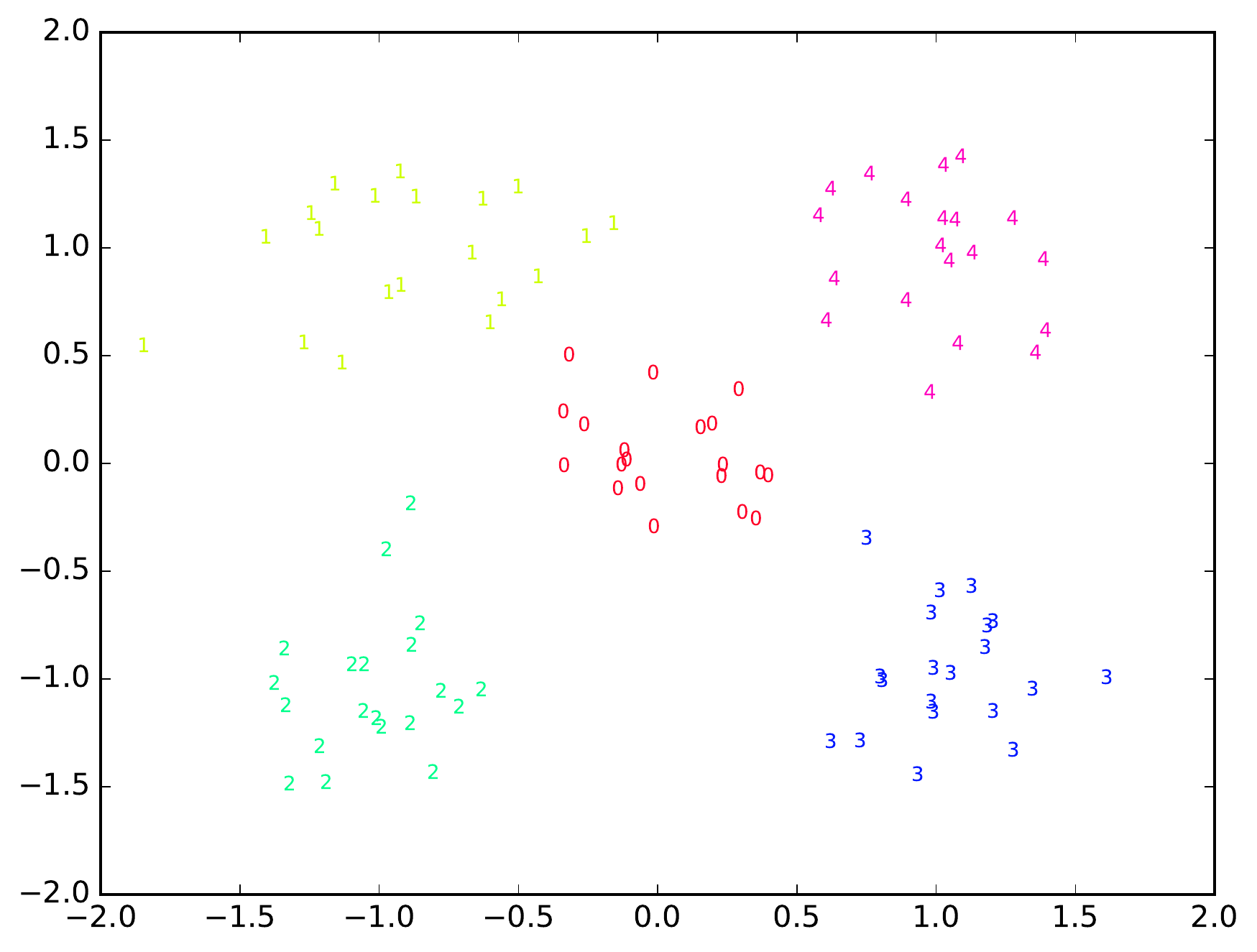}
	\includegraphics[width=0.49\textwidth]{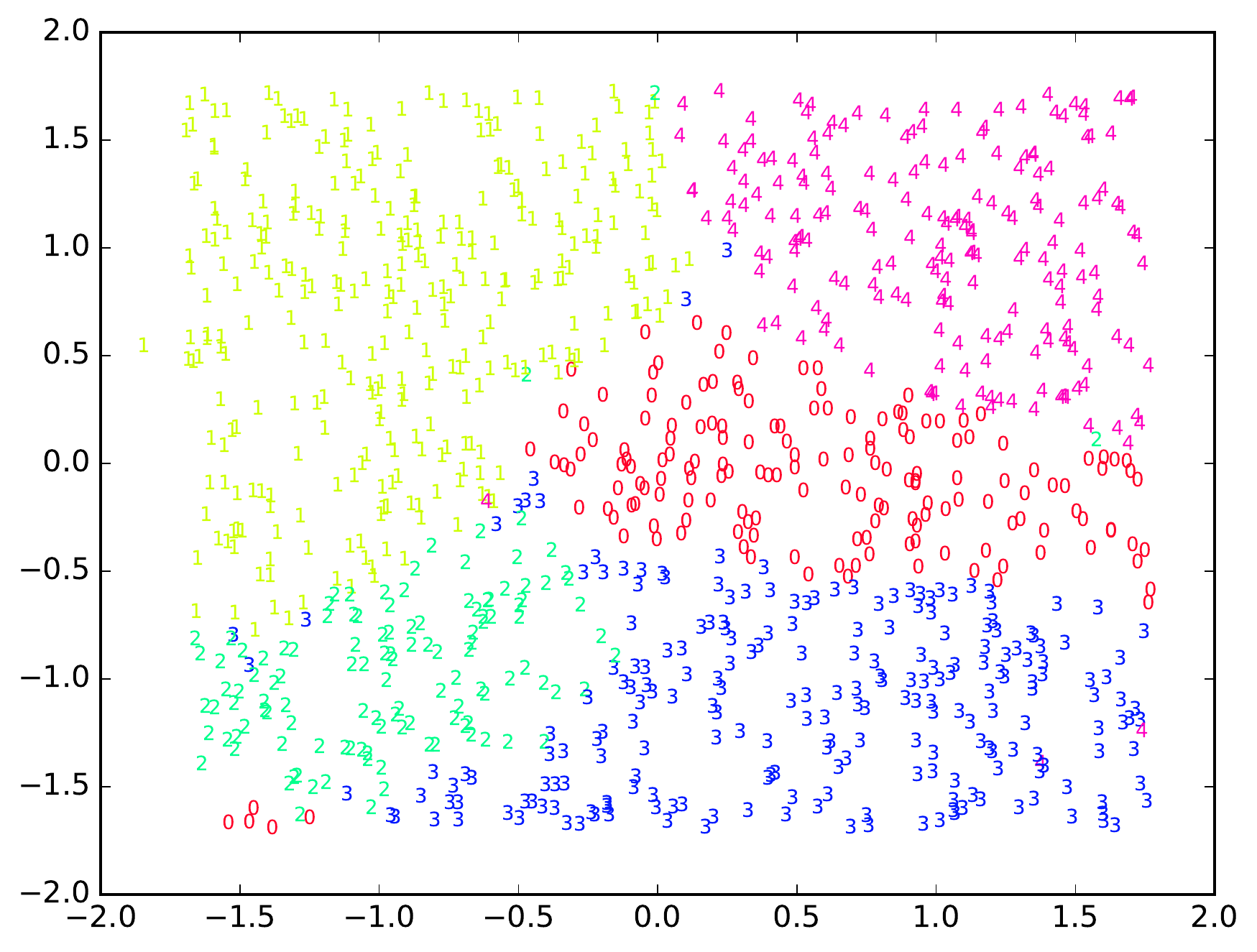}
	\caption{Semi-supervised incremental ELMVIS initialized with 20 samples per class (\emph{left}). An ELM learns all the clusters from a larger initialization set, and add new samples for all of them instead of just two (\emph{right, 1000 samples mapped}). Classes of samples are used for initialization, but ELMVIS method does not have access to them and works with raw pixel data.}
	\label{fig:ss1}
\end{figure}

In addition, a semi-supervised approach is tested where the clusters are initialized with a larger number of samples. This experiment uses 5 classes of digits with 1000 samples per class, initialized with 20 samples per class as shown on Figure~\ref{fig:ss1} (left). A large initialization set forces ELM to learn all the classes, and add samples from all of them instead of only two (Figure~\ref{fig:ss1}, right).

\begin{figure}
	\centering 
	\includegraphics[width=0.49\textwidth]{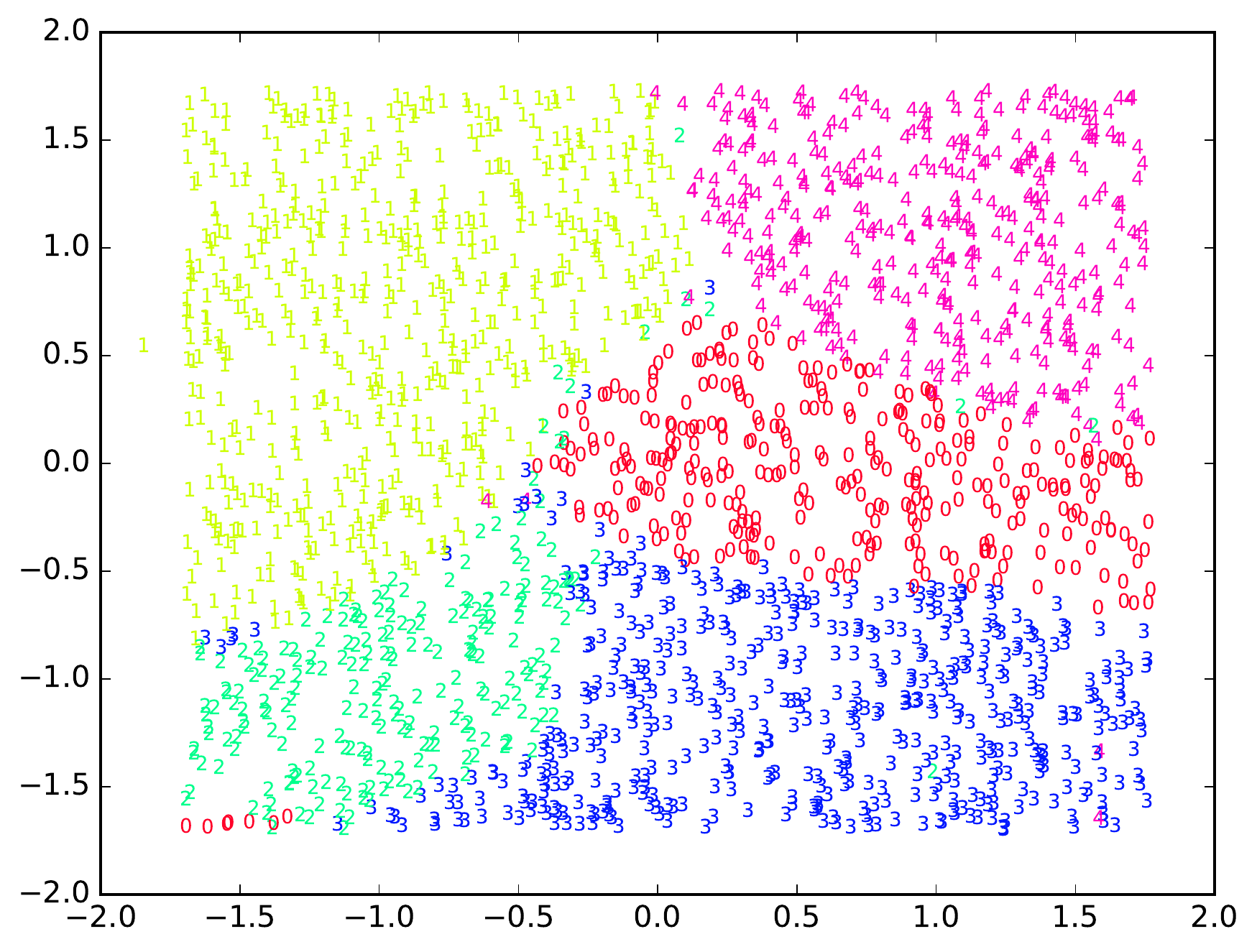}
	\includegraphics[width=0.49\textwidth]{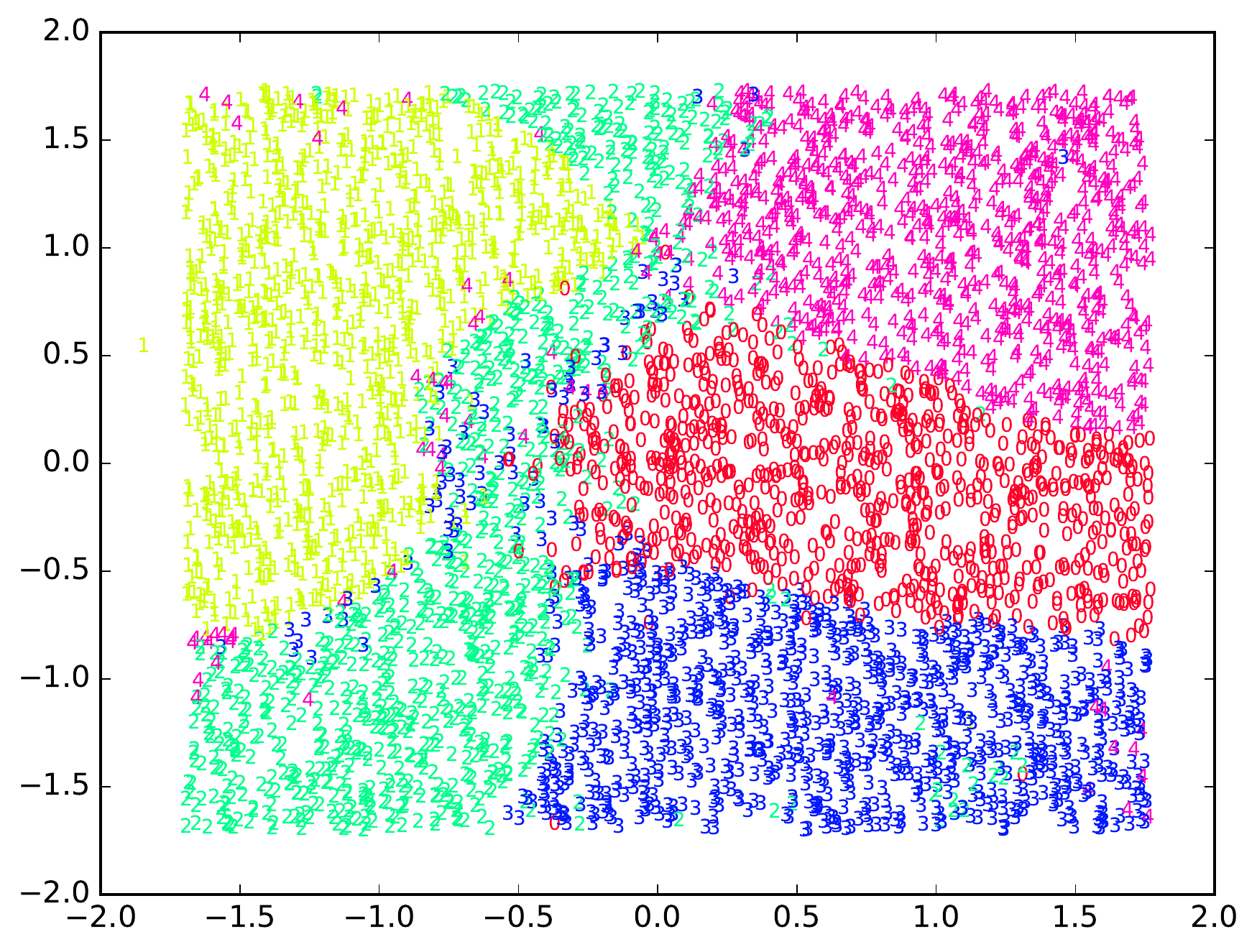}
	\caption{An additional ELMVIS+ step moves existing clusters if more space is needed for the remaining data, without disturbing the global data structure. Here 2000 samples visualized on the left with under-represented class \emph{2}; 5000 samples visualized on the right with other clusters moved to fit class \emph{2} without overlap.}
	\label{fig:ss2}
\end{figure}

When the method need space to map more samples from a particular class, an additional ELMVIS+ step moves existing clusters to give that space while keeping sharp boundaries between the classes. The effect is shown on Figure~\ref{fig:ss2} where all clusters are moved to give more space for digits \emph{2}. The global structure is well preserved, with large clusters keeping their place.

\begin{figure}
	\centering 
	\includegraphics[width=0.49\textwidth]{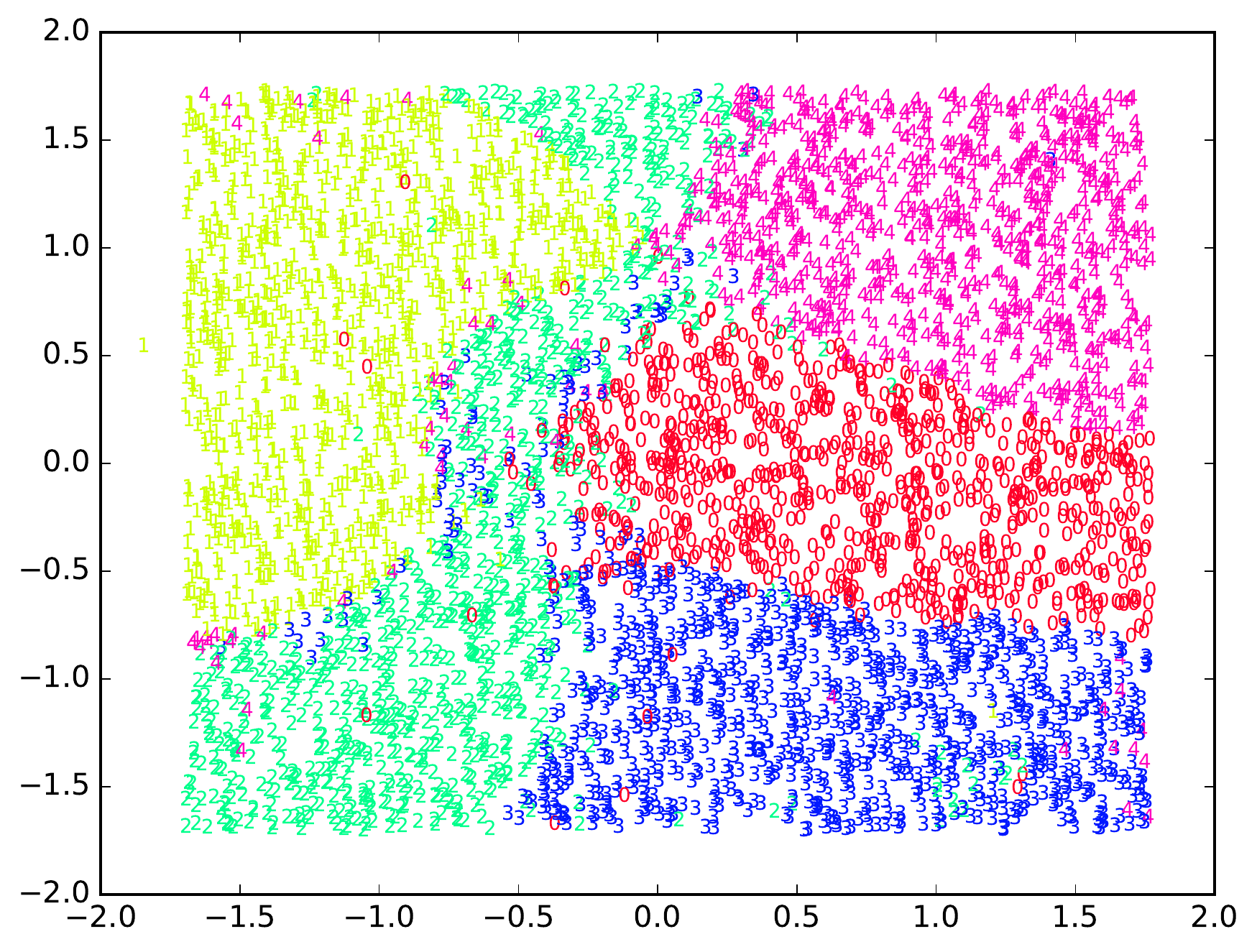}
	\includegraphics[width=0.49\textwidth]{img/ss_5000b.pdf}
	\caption{The last step on an incremental ELMVIS (\emph{left}), further refined by ELMVIS+ (\emph{right}). Candidate samples of class \emph{0} that are mis-placed due to the lack of space inside classs \emph{0} are correctly fitted by ELMVIS+.}
	\label{fig:ss_ev}
\end{figure}

The added ELMVIS+ step refines the candidate samples placement within the fixed ones, that is necessary towards the end of visualization when there may be no samples of the desired class left, or no spaces left within the desired class area. An improvement of ELMVIS+ step fitting the candidate samples is presented on Figure~\ref{fig:ss_ev}.

\subsection{Data Structure Discovery with Unsupervised ELMVIS}

The inputs to ELMVIS are not limited to visualization coordinates; they can be arbitrary data. Thus ELMVIS is a feasible method for finding structure in the data in an unsupervised manner. This experiments takes a number of MNIST digits in random (undefined) order, and maps them to the same number of classes (in zero/one encoding), or to the same number of different MNIST digits of the same classes.

First dataset has zero-one classes as inputs, and MNIST digits as outpus. However it cannot be used to train a supervised model because it is unknown which input corresponds to which output -- a common situation in the analysis of large automatically acquired data corpora that has not been manually labeled. The goal of an incremental ELMVIS is to reconstruct the correct input-output pairing. The experiment uses 100 samples per class with 2 or 5 classes of digits. Note that the best pairing across all permutations of classes is reported, as in the unsupervised setup with equal amount of samples there is no way to tell ELMVIS which class should go to which digit.

\begin{figure}
	\centering 
	\includegraphics[width=0.49\textwidth]{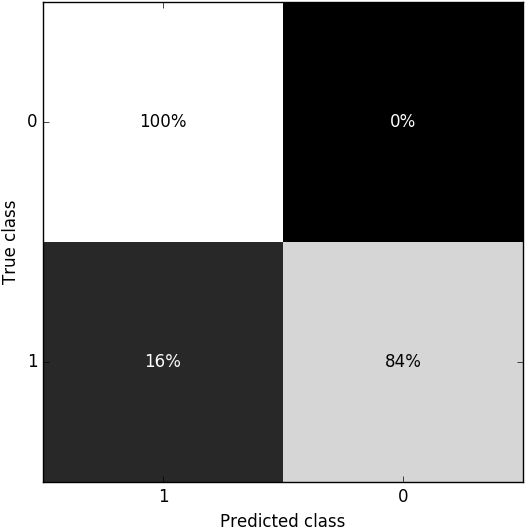}
	\includegraphics[width=0.49\textwidth]{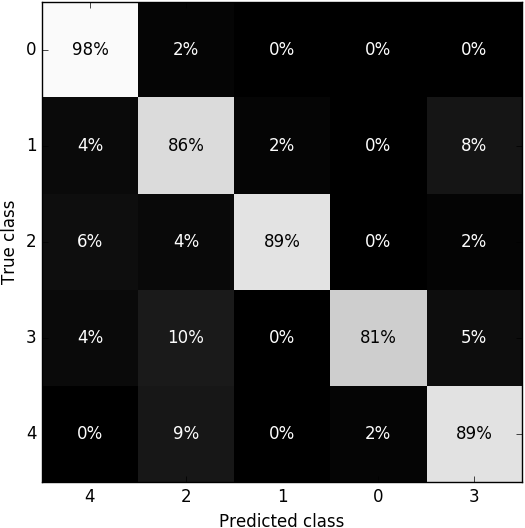}
	\caption{Binary class to MNIST digit mapping with two or five classes, confusion matrices present correct percentages. The best mapping across all permutation of classes is shown, because it is not possible to specify matching of particular classes in an unsupervised method.}
	\label{fig:match_classes}
\end{figure}

The resulting confusion matrices are shown on Figure~\ref{fig:match_classes}. An incremental ELMVIS successfully paired classes with pictures of MNIST digits. The method mapped classes arbitrary (i.e. class $[1,0,0,0,0]$ is mapped to digit \emph{4}), but this is to be expected from a purely unsupervised method.

Another experiment is performed in a similar setup, but instead of binary class representations the incremental ELMVIS tried to map MNIST digits to other MNIST digits (of the same classes). The mapping is successful with two classes. With more than two classes the same feature always appears: two random classes are separated well while other classes are randomly mixed with them. This outcome is in line with the results observed in section~\ref{sec:evbasic} where two classes are clearly separated at the beginning, followed by other classes mapped over them.

\begin{figure}
	\centering 
	\includegraphics[width=0.49\textwidth]{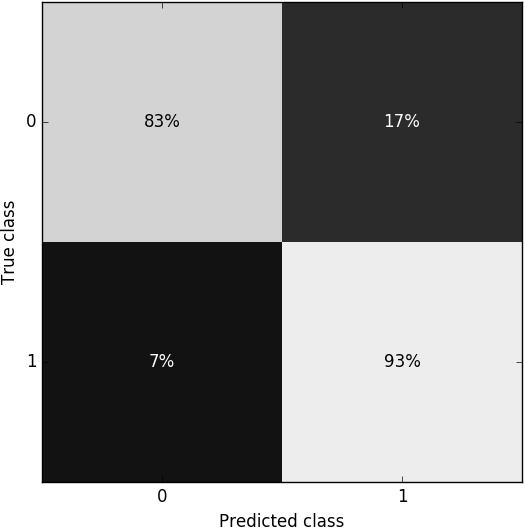}
	\includegraphics[width=0.49\textwidth]{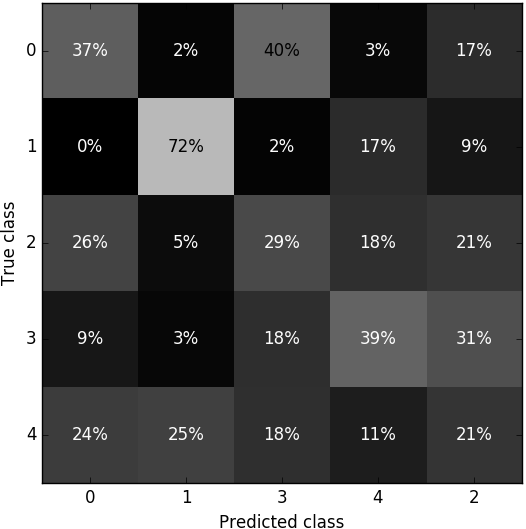}
	\caption{MNIST digits to MNIST digits mapping, using different digits from the same 2 or 5 classes. Two classes are separated clearly, while any additional number of classes are mixed with them.}
	\label{fig:match_digits}
\end{figure}

\section{Conclusions}

An iterative extension to the original ELMVIS+ method is proposed in this paper. It iteratively selects a small number of best fitting samples from all the available ones, and adds them to the model. It allows for a much larger set of potential samples than ELMVIS+ by limiting memory requirements to already fitted samples rather than to all available ones, keeping the high speed of the ELMVIS+ at the same time.

The method improves global structure of ELMVIS+ visualization by starting with a small dataset, and gradually adding more data or increasing the complexity of the model. It preserves the global structure, sharp boundaries between classes, and has a possible semi-supervised extension where the samples are mapped to the specified places on the visualization space.

The proposed method is capable of unsupervised data structure detection. It excels in reconstructing a randomly shuffled dataset with unknown pairing between inputs and outputs. It is also capable of finding a mapping between two complex data spaces as shown on MNIST digits example.

The methodology needs further investigation and improvement to counter the observed drawbacks, specifically the tendency of learning an easy model first leading to problems in incorporating more complex parts to the global picture.

\bibliographystyle{plain}
\bibliography{ds_match}

\end{document}